\documentclass[12pt, conference]{IEEEtran}
\ifCLASSINFOpdf
\else
\fi
\usepackage{graphicx}
\usepackage{caption}
\begin{document}
%
\title{Advanced Baseline for 3D Human Pose Estimation:\\ A Two-Stage Approach}

\author{\IEEEauthorblockN{Gavin (Zichen) Gui}
\IEEEauthorblockA{Master of Computing Science\\
With Specialization in Multimedia\\
University of Alberta\\
Email: zgui@ualberta.ca\\
CCID: zgui}
\and
\IEEEauthorblockN{Colton (Jungang) Luo}
\IEEEauthorblockA{Master of Computing Science\\
With Specialization in Multimedia\\
University of Alberta\\
Email: jungang@ualberta.ca\\
CCID: jungang}}


%


\maketitle

\begin{abstract}
Human pose estimation has been widely applied in various industries. While recent decades have witnessed the introduction of many advanced two-dimensional (2D) human pose estimation solutions, three-dimensional (3D) human pose estimation is still an active research field in computer vision. Generally speaking, 3D human pose estimation methods can be divided into two categories: single-stage and two-stage. In this paper, we focused on the 2D-to-3D lifting process in the two-stage methods and proposed a more advanced baseline model for 3D human pose estimation, based on the existing solutions. Our improvements include optimization of machine learning models and multiple parameters, as well as introduction of a weighted loss to the training model. Finally, we used the Human3.6M benchmark to test the final performance and it did produce satisfactory results.
\end{abstract}


%
\IEEEpeerreviewmaketitle

\section{Introduction}
As part of the ongoing research in computer vision, human pose estimation is the process of extracting and constructing transformations and locations of human joints from images or videos. In recent decades, many traditional industries are utilizing this technique to benefit customers, including video surveillance, medication, education, robotics and so on \cite{WANG2021103225}. Therefore, it has become a popular computer-vision-based research area and many previous studies have presented different solutions. 

For example, a classical approach used to estimate the 2D human poses is called the pictorial structure framework (PSF) \cite{6619235}. It uses a discriminator to identify body parts and a prior to model the probability distribution over pose. However, due to its inability to detect and construct body parts hidden from a certain perspective or occlusions, such a classical approach was destined to be abandoned for better accuracy. Nowadays, people are turning to the latest and more state-of-the-art techniques such as deep learning, and convolutional neural networks (CNN) is one of the approaches that have been most widely adopted.

On the other hand, 3D human pose estimation is the process of extracting articulated 3D joints' coordinates. Existing 3D human pose estimation methods can be commonly categorised into two types: single-stage and two-stage \cite{https://doi.org/10.48550/arxiv.2012.13392}. The first type \cite{Li20143DHP} \cite{https://doi.org/10.48550/arxiv.1611.07828} obtains 3D key points’ positions directly from image regression or other techniques. Some early single-stage methods even directly used some deep learning models to establish the end-to-end mappings between the RGB images and the 3D results without any intermediate steps, but the biggest problem of these straightforward solutions is that such mappings are highly non-linear and may lead to inaccuracy caused by overfitting. Recent development also proposed other advanced single-stage models, such as the Decoupled Regression Model \cite{9879368} where the researchers adopted new decoupled 3D pose representation and advanced 2D pose extraction module.

Compared with the first one, the second type \cite{https://doi.org/10.48550/arxiv.1705.03098} \cite{https://doi.org/10.48550/arxiv.1612.06524} \cite{https://doi.org/10.48550/arxiv.2008.00206} uses various 2D human pose estimation as an extra intermediate stage and then lifts 2D results to 3D poses as the second stage. The most significant advantage of these methods is the lightweight network and faster processing. However, they are highly dependent on the 2D human pose estimation process and any error or mistake in 2D may become distinctly noticeable in 3D. But as a result of the recent rapid development, 2D estimation has been quite advanced and the error should be acceptable in most situations. Another potential problem of most two-stage methods is that some information in the original RGB image (e.g., spatial data) can be lost after the first stage, which may lead to some inaccuracy in the second stage. For example, in \cite{https://doi.org/10.48550/arxiv.1612.06524} the researchers built a large 2D-3D pose pair library so that they can use the 2D pose to match a most close 3D pose. Though it does work, some depth data can be lost during this process.

In the following sections of this paper, we are going to propose a new advanced baseline model for 3D human pose estimation and our research focuses on the 2D-to-3D lifting process of the two-stage approach. We will first look into and critically review the state-of-the-art researches in the computer vision field as well as the human pose estimation field. Then, we will elaborate on our proposed improved model in detail and quantitatively compare ours with the existing ones. Finally, we will discuss the results and reach a conclusion.

\section{Literature Review}
\subsection{Publications of the Multimedia Research Centre}
A great amount of research has been conducted in the fields of computer vision and multimedia within decades. For example, since the 1990s much attention has been paid to teleconferencing, the technology which allows users to use the Internet to visually communicate with other people. Considering the limited network bandwidth and computer performance in the twentieth century, \cite{390703} and \cite{661143} proposed a new lossy image compression method called variable resolution transform. This transform is aimed at keeping more details in an area of interest (fovea), while reducing details in the remainder (periphery) to compress data. Following that, an inverse transform will be applied when image decompression is desired. 

To be more exact, while \cite{390703} can be viewed as a brief introduction of  the variable resolution model, \cite{661143} provides deeper insights into the improved transform and proposed a combination of this model and the JPEG (Joint Photographic Expert Group) scheme to further optimize. Regardless, both research results showed that significant overheads could be reduced as only a reduced part of the image need to be transmitted and further compressed. In addition, these two papers also introduced the possibility of movable and multiple foveas, as well as three new interpolation methods for improvement. Since video conferencing did not demand high quality at that time, the researchers concluded that this lossy compression method is indeed able to provide a good teleconference experience. However, there exists one potential problem which the authors did not attempt to address: when the local sampling rate is very low, variable resolution may lead to aliasing and more noise on the image.

Similarly, limited transmission capacity also has a negative impact on telepresence. Thus, in the paper \cite{770389} the authors presented a new solution by attempting to apply the predictive Kalman filter to predict which part of the image would be required next, and then display panoramic videos captured by their proposed system to the human operators. Quantitatively, their experiments did produce better results than standard methods without prediction, proving its advantages of reduced delay and higher refresh rates. Therefore, this research can inspire and promote similar applications of prediction models even today, and promising results can be expected.

Apart from the quantitative measures of data transmission over the network, \cite{4066986} raised a commonly-neglected issue, namely human’s perception. The authors first presented a framework for human's perception estimation and then analyzed five potential solutions based on this framework. Finally, the best one was chosen as the proposed new method to produce better human perception over a weak network. However, as mentioned in the authors’ conclusion, the packet loss model adopted in the experiments is preliminary and thus the actual weak network circumstances may see slightly worse results. Regardless, this paper still took a novel step and inspires future studies to take human factors into consideration when introducing a new technology.

These early studies discussed above have laid a foundation for modern multimedia advancements. Entering the twenty-first century, another topic that has been heatedly discussed is human recognition. Among the existing substantial research in such distinctive field, \cite{1594295}, \cite{1545408} and \cite{941210} proposed innovative methods to respectively tackle the problems of pose recognition, hand gesture recognition and nose shape recognition. 

In particular, \cite{1594295} and \cite{1545408} both utilized the parameterized Radon transform, which requires no normalization of the image and less computational expenses than many other template-based approaches, and promises a relatively high recognition rate. They first compute a binary skeleton representation of the human subject. Prior to using K-means to classify different gestures, \cite{1594295} uses the spatial maxima mapping algorithm to match unknown poses and known poses, while \cite{1545408} performs the extraction of the most important features by computing the normalized cumulative projections of Radon transform. However, as \cite{1545408} pointed out, one disadvantage of applying K-means here is that an extra step must be taken to compute the known-pose dataset's nearest neighbours of the centroids of the unknown poses.

Another paper \cite{941210} has a relatively high recognition rate in detecting noses. It used a twisted pair curve template to resemble the nostril shape and two symmetrically paired vertical-parabola-like templates to resemble the shape of nose sides, and the cost function was defined to estimate the brightness and edge magnitude. The results proved the approach feasible in practical cases, whereas there are some issues to be addressed. For example, the authors admitted that this approach may not apply to cases when the nose orientation is not parallel to the face centre line, or when the face rotates too much. Nevertheless, this research made an important first move to include nose detection as part of facial detection, which may contribute to future demands and applications and encourage further studies on or based on nose detection.

\subsection{Publications on the SmartMultimedia Conference}
With the recent development of Light Detection and Ranging (LiDAR) devices, people can now easily generate 3D point clouds, which can be useful in various real-world industries. One example is the new method to estimate the buttress volume of trees introduced in \cite{10.1007/978-3-030-54407-2_39}. Based on the LiDAR devices and the Poisson Surface Reconstruction algorithm, the authors achieved an accuracy of around 95\%. This research is generally a good preliminary application of LiDAR devices, but the designed experiments can be more persuasive and it only combined two existing methods without any noticeably fundamental innovation.

Once the point clouds have been generated, they may be aligned to actual objects for registration or navigation, and \cite{10.1007/978-3-030-54407-2_14} discussed a new unsupervised network (DPCAN) to achieve such point-cloud alignments. It first maps unordered point clouds onto a 2D space so that it is able to increase efficiency by applying a linear classifier. Then, it estimates the object’s pose by finding two point-cloud pairs based on the least squares method. Overall, this research can be seen as an effective solution to point-could alignment and its recognition accuracy is more than 80\%. Its success also reveals that some difficult challenges may be able to be simplified by techniques such as dimension reduction.

Classification of these 3D data has also been frequently discussed in recent decades. For example, \cite{10.1007/978-3-030-54407-2_38} proposed a new simple classification technique where multiple 2D perspectives of 3D point clouds are used as input, meanwhile most previous attempts used only one perspective. However, similar to \cite{10.1007/978-3-030-54407-2_39}, the following stages of their research highly depended on the existing tool (YOLO v3). Also, though their model did produce a high accuracy of over 98\%, the researchers may consider including more information from the point clouds (depth, etc.) to further enhance their preprocessing stage.

Apart from the 3D point clouds, another interesting topic among various multimedia technologies is background subtraction. It has been one of the most commonly used methods for object detection, especially in the use cases of detecting moving objects from video frames captured by a static camera. By definition, it is an approach that extracts the foreground from the background of an image for later processing. 

Traditional background subtraction techniques are usually deemed as performing pixel classification over time, which could lead to relatively long computational time and not-so-satisfying F-measure and precision. Fortunately, a recent innovative approach called Difference Clustering was proposed in \cite{10.1007/978-3-030-54407-2_4} to overcome these weaknesses as the researchers adopted the strategy of grouping into two clusters the difference vectors between current frame and the reference frame. The result of this approach proved it to be a notable option for real-time detection of moving objects in a video, with over 90\% of computational time saved compared to previous approaches. 

However, with the advancement and popularization of mobile phones and cameras, background subtraction has become more difficult to achieve, as there emerges an increasing amount of use cases when it requires detecting moving objects from a freely moving camera. Since rarely any existing approach can appropriately handle dynamic image backgrounds, \cite{10.1007/978-3-030-54407-2_6} proposed a method using Robust Principal Component Analysis (RPCA) for the background motion analysis. The method first captures the optical flows to describe the motions of pixels to generate a motion matrix, followed by utilizing RPCA to decompose the motion of the background for background subtraction. Although the generation of optical flows created some defects that will affect the accuracy, the accuracy of this method can be enhanced with the help of super-pixels. More importantly, this new method can act as an inspiration for better solutions to detecting moving objects filmed by a freely moving camera.

Background subtraction sees its application in the human body fall recognition system. In \cite{10.1007/978-3-030-54407-2_31}, different background subtraction methods such as GMG and MOG2 are utilized to extract the foreground and background frames, after which the human body fall features are detected in the processed frames. Although the final results of fall detection accuracy tests reached a relatively high percentage of 85\%, the researchers suggested it could still be improved by solving some existing issues in four aspects, namely illumination rates for compendious human body outline, non-target moving objects in the frames, camera-to-target distance and occlusion. As stated in \cite{10.1007/978-3-030-54407-2_31}, measures to solve these problems include applying other background subtraction techniques that are free from constraints of human shapes and angles to extract of certain human body features. From such a real-life application, it is apparent to see the fundamental role that well developed background subtraction techniques play.

\subsection{2D Human Pose Estimation}
As a precondition for 3D human pose estimation, 2D human pose estimation algorithms are used to calculate the location of human joints in images or videos. For traditional 2D human pose estimation algorithms, hand-craft feature extraction and sophisticated body models were required to get local representations and global pose structures until deep learning revolutionized the field. In recent years, 2D human pose estimation has been undergoing significant advances and magnificent breakthroughs as deep learning techniques have emerged. As suggested by the survey \cite{Chen_2020}, deep-learning-based 2D human pose estimation methods can be classified as detection-based or regression-based.

Derived from the body part detection methods, detection-based 2D human pose estimation methods are designed to provide an estimate of the locations of body joints or limbs. They are typically guided by a sequence of rectangular windows or heatmaps. Substantial newly published studies utilized heatmaps to describe the ground truth of the joint location in an attempt to offer additional supervision information as well as to assist in training networks. The heatmap channel for each joint in \cite{10.1007/978-3-030-01219-9_12}, for example, has a 2D Gaussian distribution centered on the joint location.

In regression-based 2D human pose estimation techniques, joint coordinates are directly derived from the image using an end-to-end framework that maps the image to kinematic body joints. A soft-argmax function was proposed in \cite{https://doi.org/10.48550/arxiv.1710.02322} to transform heatmaps to joint coordinates in order to preserve both the benefits of heatmap supervision and numerical joint positions supervision, thereby turning a detection-based network into a differentiable regression-based network. In particular, the soft-argmax operation can be incorporated into a deep convolutional network to indirectly obtain part-based detection maps. This approach yields a considerable improvement over the existing benchmark scores from regression-based methods as well as competitive outcomes over detection-based approaches. For this reason, it may be considered an attractive option for our study.

Additionally, it should be mentioned that existing methods often only focused on directly obtaining 2D joint coordinates and neglected the uncertainty associated with the 2D detector, namely the depth ambiguities and occluded body parts. But \cite{https://doi.org/10.48550/arxiv.2107.13788} specifically proposed a method to obtain and utilize the uncertainties associated with the 2D detector from the predicted heatmaps. In order to best capture uncertainty distribution, the study fits a 2D Gaussian to each predicted heatmap using nonlinear least squares. By doing so, such uncertainty as well as the inherent depth ambiguities can be accurately modelled, so as to preserve important data for 2D-to-3D lifting later. 

\subsection{3D Human Pose Estimation}
Unlike 2D pose estimation, the objective of 3D human pose estimation is to estimate the positions of human joints in a 3D coordinate system \cite{https://doi.org/10.48550/arxiv.2012.13392}. That is to say, apart from obtaining the values of the x and y axis, 3D pose estimation also requires the estimation of the z axis, or the depth of the human body. One of the most popular datasets in this area is called Human3.6M \cite{6682899}, which has been extensively used in previous studies. It includes a total of 3.6 million different human poses, made up of eleven people performing seventeen indoor scenes. There is a marker-based motion capture system in the laboratory to record the joint positions of the actors or actresses, as well as four cameras placed to capture them from different angles.

Previous researchers have indeed introduced some 3D human pose estimation solutions that can produce good results, and they can usually be divided into two categories: single-stage and two-stage. The main difference between these two categories is whether 2D human pose estimation methods and 2D-to-3D lifting processing are adopted \cite{https://doi.org/10.48550/arxiv.2012.13392}.

A classic single-stage solution is stated in \cite{Li20143DHP}, which is one of the first studies that introduced deep learning to the field of 3D human pose estimation. The researchers designed a multitask convolutional network including both body detection and pose regression, so that the proposed model can directly learn from monocular images and then generate and estimate the positions of 3D joints. As the first attempt in this field, its experimental results quantitatively proved that this model did significantly outperform traditional methods at that time. 

\begin{center}
    \includegraphics[width = .5\textwidth]{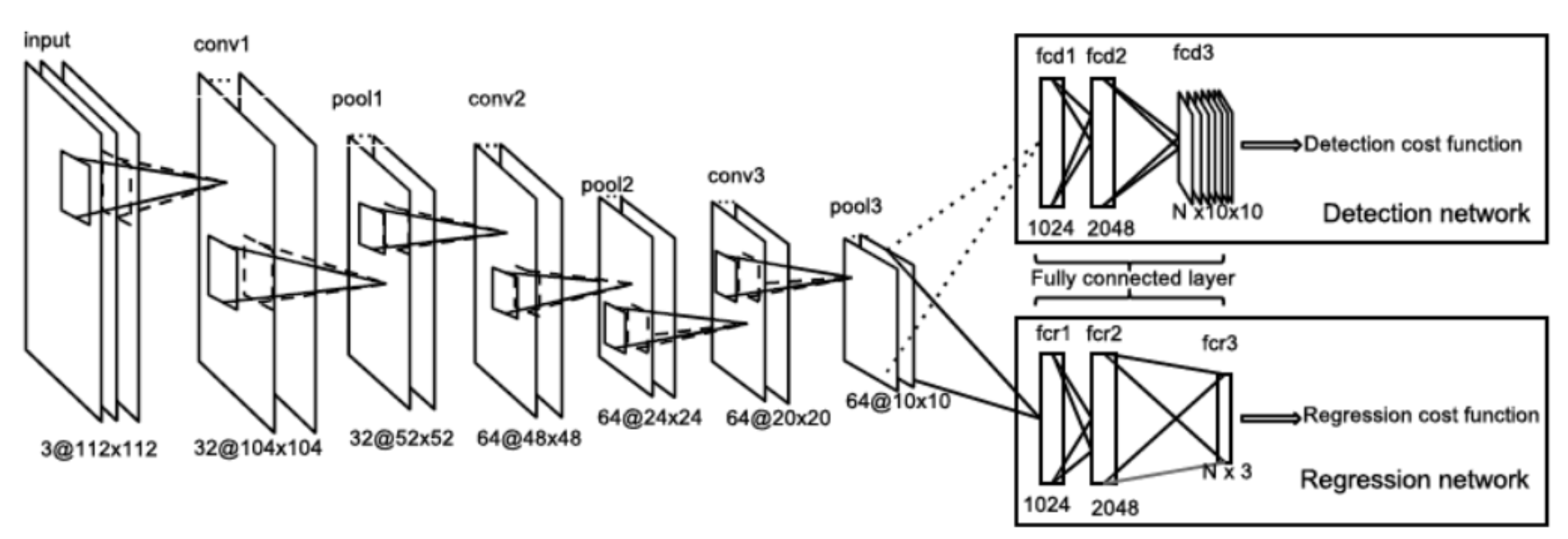}
    \captionof{figure}{Architecture of \cite{Li20143DHP}}
\end{center}

\cite{https://doi.org/10.48550/arxiv.1611.07828} is another single-stage method whose network may seem to resemble the Hourglass network in 2D estimation. It used volumetric heatmaps to describe human joints, and employed a coarse-to-fine strategy to gradually obtain precise values of the z axis. However, though according to the authors 30\% errors can be reduced on average in this model, it should be mentioned that the volumetric representations will inevitably consume much more computational resources and takes up more space.

\begin{center}
    \includegraphics[width = .5\textwidth]{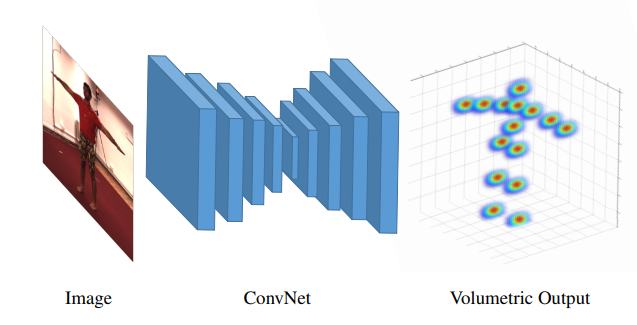}
    \captionof{figure}{Architecture of \cite{https://doi.org/10.48550/arxiv.1611.07828}}
\end{center}

Such single-stage estimation methods can better utilize information from the original image as all the data will be preserved throughout the learning networks, while they can be lost in the transitions of two-stage models. However, the major problem of single-stage models is that direct mapping between 2D images and 3D poses is highly non-linear as the projection of different 3D poses to 2D images can be the same, which may lead to more potential inaccuracies. Therefore, many researchers also turned to two-stage models to seek improvements.

An early attempt of the two-stage method is elaborated in \cite{https://doi.org/10.48550/arxiv.1612.06524}. As the paper name suggests, \textit{3D Human Pose Estimation = 2D Pose Estimation + Matching}, once given the 2D pose input, this study simply used the nearest neighbour algorithm to find the most similar 3D pose in the training data. That is to say, it first projected all the 3D poses into 2D space and simply compare the input 2D pose to find the nearest one as output. Generally speaking, such a simple matching algorithm can work well if the training dataset is very large and representative. However, in actual cases it can be less precise and suffer from noticeably errors.

\begin{center}
    \includegraphics[width = .3\textwidth]{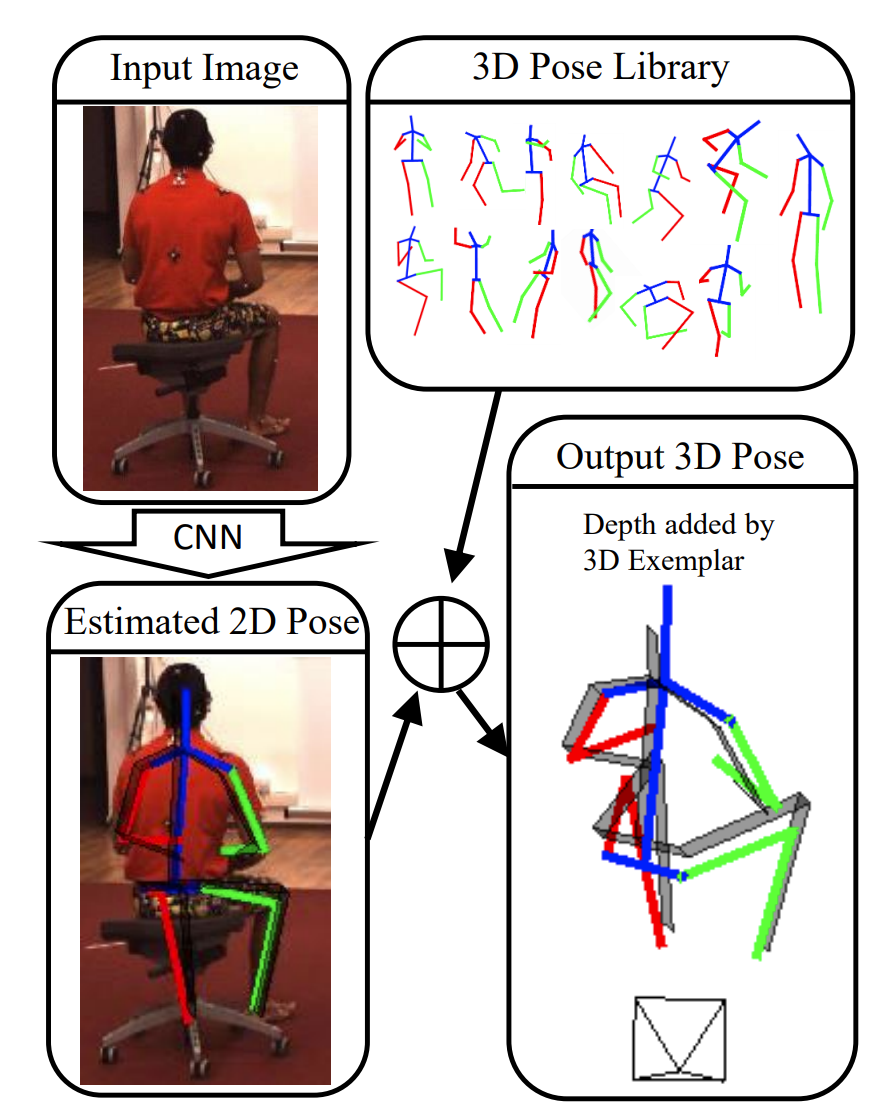}
    
    \captionof{figure}{Architecture of \cite{https://doi.org/10.48550/arxiv.1612.06524}}
\end{center}

\cite{https://doi.org/10.48550/arxiv.1705.03098} proposed a very famous yet very straightforward two-stage baseline for 3d pose estimation. It mainly focused on the design of the 2D-to-3D lifting process, which takes 2D positions as input and then used residual connections and batch normalization techniques to estimate 3D positions as output. Compared to the last one, \cite{https://doi.org/10.48550/arxiv.1705.03098} achieved much better results with remarkably low error rates. This research also indicated that the errors in modern two-stage methods are mainly due to the problematic 2D estimation, rather than the lifting process. This finding further justified two-stage approaches and paved the way for future advancements in this direction.

\begin{center}
    \includegraphics[width = .5\textwidth]{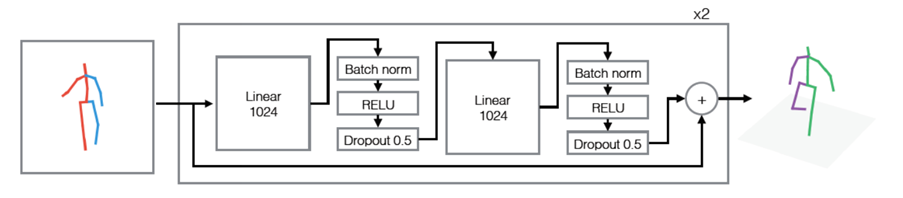}
    \captionof{figure}{Architecture of \cite{https://doi.org/10.48550/arxiv.1705.03098}}
\end{center}

\cite{https://doi.org/10.48550/arxiv.1902.09868} is a more recent and advanced two-stage approach and uses a weakly supervised model. As shown in the figure below, this model includes three main components: a pose and camera network which estimates 3D pose and camera matrix from 2D coordinates, a critic network which determines whether the 3D pose produced is valid and realistic, and a reprojection network to convert 3D pose back to 2D pose and do the weakly supervised learning. The experiment results showed that this model can generalize well and achieved a higher Human3.6M benchmark score over the state-of-the-art methods at that time.

\begin{center}
    \includegraphics[width = .5\textwidth]{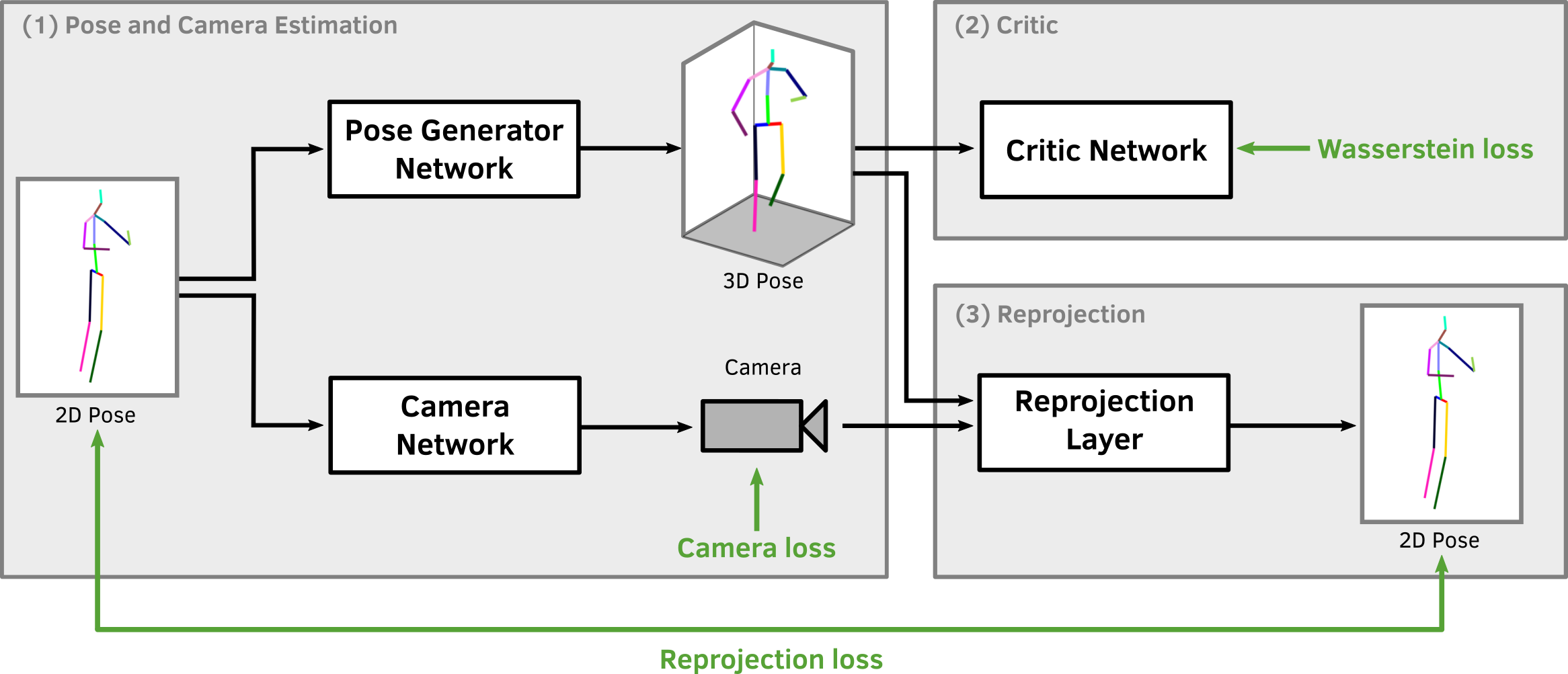}
    \captionof{figure}{Architecture of RepNet \cite{https://doi.org/10.48550/arxiv.1902.09868}}
\end{center}

\section{Methods}
\subsection{Base Model}
Due to the limited research time span, instead of building everything from scratch, our proposed model is based on the 3D human pose estimation baseline \cite{https://doi.org/10.48550/arxiv.1705.03098}. Though it has been around for five years since its publication, this lightweight model is still widely utilized today. In addition, considering the simplicity of this baseline, its training process does not take a significantly long time. Therefore, we can make changes to it and see the results in time so that we can have multiple iterations within such short time to gradually improve our proposed model.

\subsection{Amelioration and Refinements}
Throughout this research, our team have carried out a significant number of different experiments and trials, and achieved a few successful iterations. They are summarized into the following three major versions that we are going to eloborate in detail.
\subsubsection{Version 1}
As the first attempt to improve the results of the model, an increased number of linear and batch normalization layers are added to the model. A linear layer in a neural network performs on the input data a linear operation, which typically involves multiplying the input by a weight matrix and adding a bias term. The output of a linear layer is often passed through a non-linear activation function, such as the Rectified Linear Unit (ReLU) function in this case, to introduce non-linearity into the network. Batch normalization, on the other hand, is a technique used to normalize the inputs of a layer in a neural network, which typically involves normalizing the inputs to have zero mean and unit variance. This can help the network to learn faster and reduce the chances of getting stuck in a local minimum, so as to enhance the performance of the model. By increasing the number of these two layers to the existing base model, we expect to see slightly improved results over the base model. Meanwhile, since we do not want to make the model over-complicated, we simply added one more linear and batch normalization layer to the model to keep it lightweight.

\subsubsection{Version 2}
Our second investigation towards finer results is launched through replacing the activation function ReLU with Swish, despite the extreme high usage of ReLU in the recent decade. Swish is a more recent activation function that was developed by the Google Brain Team \cite{https://doi.org/10.48550/arxiv.1710.05941}. According to them, Swish is simple and similiar to ReLU, but it has been proved to outperform ReLU on a variety of deep learning models. Taking it into consideration, we decided to use it in our model.

The Swish activation function is described as follows, where $\beta$ is a learnable parameter:
\begin{center}
    \includegraphics[width = .3\textwidth]{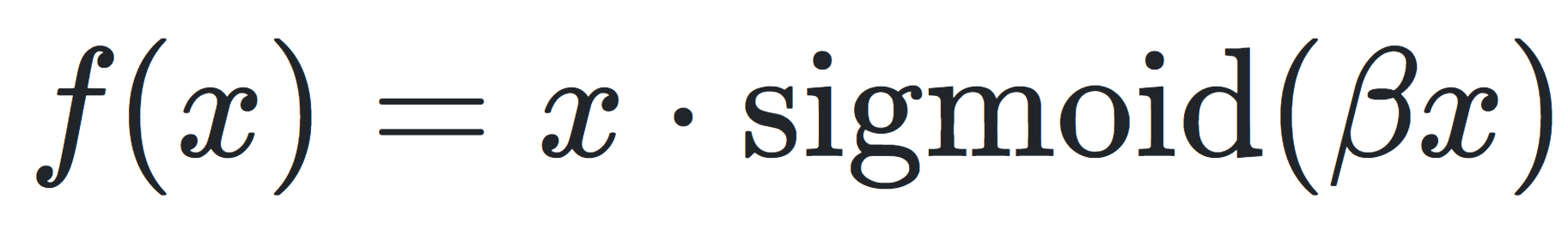}
    
    \captionof{figure}{Formula of the Swish Function \cite{https://doi.org/10.48550/arxiv.1710.05941}}
\end{center}
In addition to that, we have also optimized multiple training parameters in hope to achieve more satisfactory results, including but not limited to linear size, batch size, loss function (L1 or L2 Loss), dropout rate and so on.

By the end of Version 2, our improved network architecture can be summarized in the following figure.
\begin{center}
    \includegraphics[width = .5\textwidth]{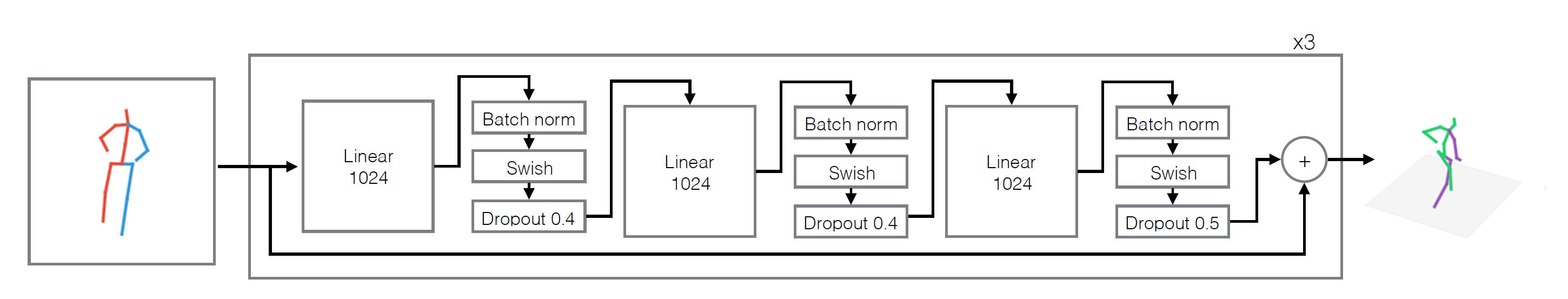}
    \captionof{figure}{Proposed Improved Architecture}
\end{center}
\subsubsection{Version 3}
The third version included our greatest novelty in this project, which is to add weighted loss to the model so that the predicted results can be more useful in practice. Despite the undeniable importance of all the human joints to be predicted, in our actual usage, some of them can be even more crucial than the others. For example, if we compare the spine with the left wrist, in most cases we can say that the estimation of the left wrist can be somehow more trivial here. Therefore, we decided to give them different weights to better fit in the real-world usage. 
\begin{center}
    \includegraphics[width = .5\textwidth]{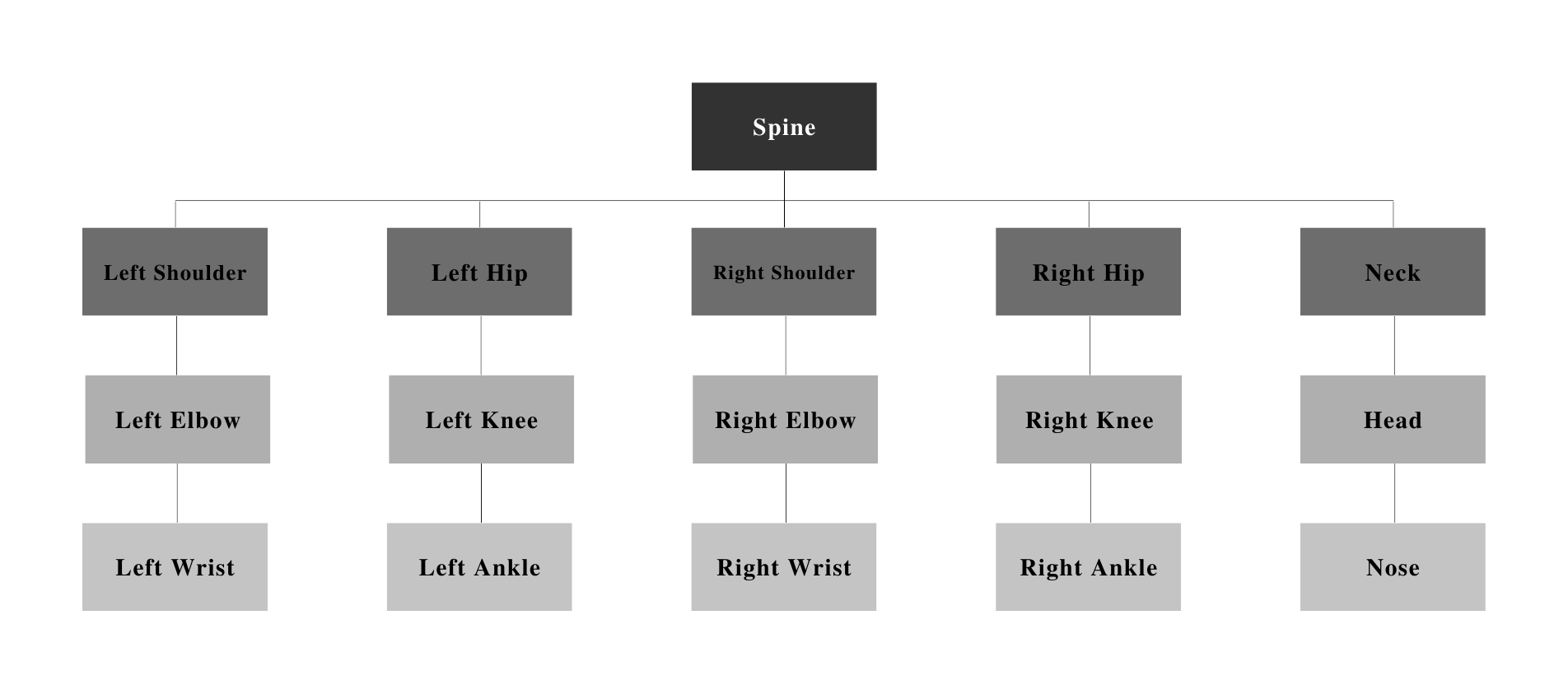}
    \captionof{figure}{Proposed Importance Rank of Human Joints}
\end{center}

As a preliminary attempt to introduce weighted joint loss, in this study we simply gave them weights of 4, 3, 2, and 1, respectively, and used the Weighted Mean Squared Error (WMSE) as the final loss function. The formula of WMSE is shown as follows, where $Wi$ is the designed weight.
\begin{center}
    \includegraphics[width = .4\textwidth]{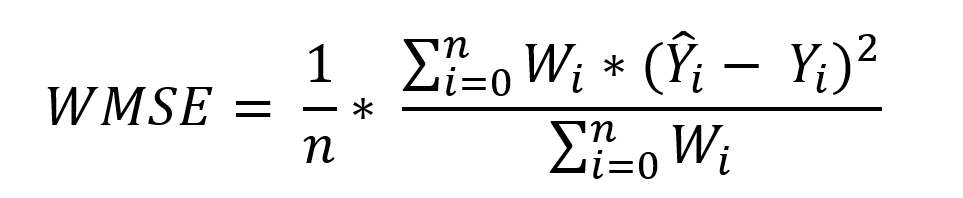}
    
    \captionof{figure}{Formula of WMSE}
\end{center}

\subsection{Model Training}
As previously explained, we used the Human3.6M as our dataset. It includes 3.6 million different human poses data and they are divided into 7 subsets. As a deep learning algorithm, we used the first 5 subsets for model training, and the remaining 2 subsets for model testing. To keep a balance between the training time cost and the loss, we slightly decreased the number of epochs from 200 to 150. Our experiment results showed that it did not result in noticeably higher errors.

We implemented our advanced model based on the PyTorch implementation of the baseline \cite{https://doi.org/10.48550/arxiv.1705.03098}. It took us approximately 5.5 minutes for each training epoch on a Nvidia 3050 Ti Graphic Card, and thus around 14 hours to train the whole model for 150 epochs. Such time cost is acceptable to us, which allowed us to gradually improve our CNN architecture and gave us adequate time to re-train it for multiple times.

\section{Results}
To evaluate our results quantitatively against the original method, we used the Mean Per Joint Position Error (MPJPE) as a testing metric for performances on the latter two subsets of the Human3.6M dataset. The MPJPE is calculated by finding the Euclidean distance between the predicted and ground truth positions for each joint, and then taking the mean of these values for all 16 joints. A lower MPJPE value indicates a better result.

As can be seen in the Table 1, the performance of the model was indeed improved by merely adding more linear and batch normalization layers to the base model. There were some specific poses where version 1 performed particularly well, including but not limited to $Phoning (phone)$, $Purchases (purch.)$, and $Sitting Down (sitd.)$ poses, with the $Sitting Down (sitd.)$ pose having the most substantial reduction among all, from 58.0 to 53.9. However, the only downside of the version lies in the slight increase from 40.3 by 0.1 in the MPJPE value of the $Eating (eat.)$ pose.
\begin{center}
    \includegraphics[width = .5\textwidth]{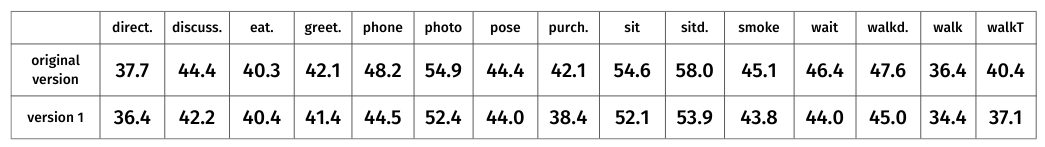}
    \captionof{table}{Result Comparison Between Version 1 and Original Version}
\end{center}

Table 2 presents a comparison of the results after changing the activation function from ReLU to Swish and tuning certain parameters. As the differences in the MPJPE values tell, version 2 had overall more significant improvements than version 1 did compared to the original version. To be specific, the MPJPE value of the $Phoning (phone)$ pose dropped drastically from that of the original (from 48.2 to 43.8), while that of the $Posing (pose)$ pose was reduced by the most amount from that of version 1 (from 44.4 to 41.7). Notwithstanding, version 2 performed badly in such poses as $Purchases (purch.)$ (worse than the original), $Sitting Down (sitd.)$ and $WalkTogether (walkT)$ (both worse than version 1).

\begin{center}
    \includegraphics[width = .5\textwidth]{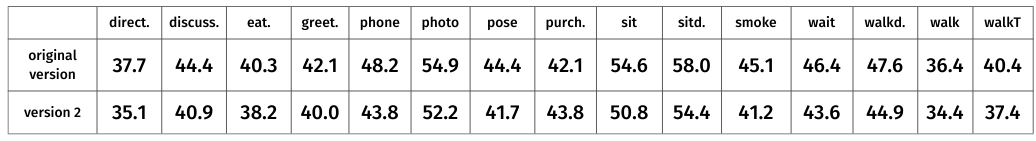}
    \captionof{table}{Result Comparison Between Version 2 and Original Version}
\end{center}

As is shown in Table 3, in spite of the result not being as appalling as the previous two modified versions, the introduction of weighted loss function to the model still pushed the performance to a higher ground compared to that of the original model. Version 3 also had a notable improvement in the MPJPE values for the $Discussion (discuss)$ and $Sitting Down (sitd.)$ poses.

\begin{center}
    \includegraphics[width = .5\textwidth]{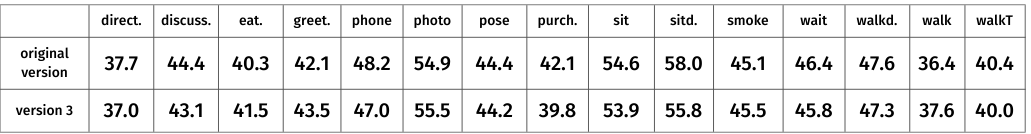}
    \captionof{table}{Result Comparison Between Version 3 and Original Version}
\end{center}

To summarize the above results, it can be seen that version 1 and version 2 of the method performed better than the original version in terms of the MPJPE metric, with an average improvement of 6.6\% and 12.2\%, respectively. In contrast, version 3 did not perform as well as the original version, with only a 1\% increase in the MPJPE. We are going to further discuss the results and reasons behind them in the next section of this paper.

\begin{center}
    \includegraphics[width = .5\textwidth]{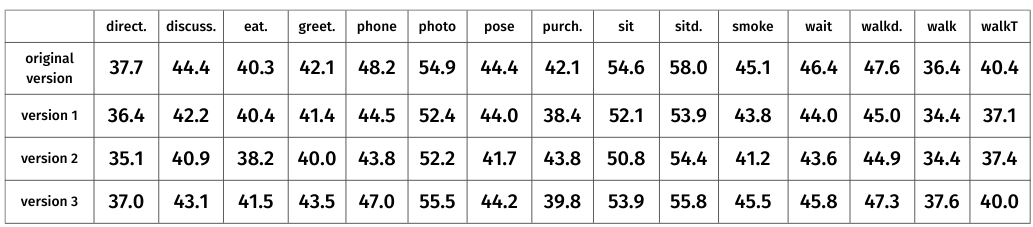}
    \captionof{table}{Result Comparison with All Versions}
\end{center}

Figures 10 to 14 below demonstrate some of the persuasive visualized results from the final tuned model (version 3 above). From left to right in each figure are respectively the 2D Pose Ground Truth, 3D Pose Ground Truth and Predicted 3D Pose. It is apparent that, even with the seemingly underwhelming MPJPE values as of the current moment, the major joints (e.g., spine) of the poses are estimated by version 3 in quite a high accuracy, and all the observable minor prediction errors only happened to the less important articulations which were purposely assigned with lower weights.
\begin{center}
    \includegraphics[width = .5\textwidth]{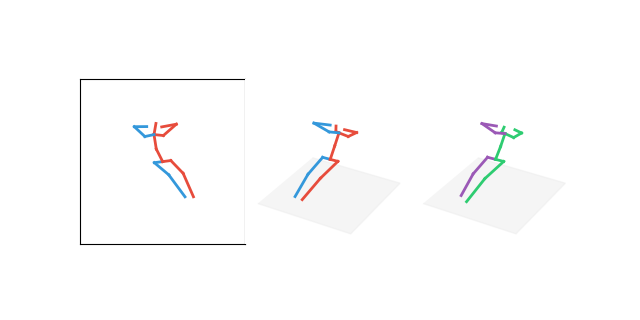}  
    \captionof{figure}{Sitting Pose Estimation Result}
\end{center}

\begin{center}
    \includegraphics[width = .5\textwidth]{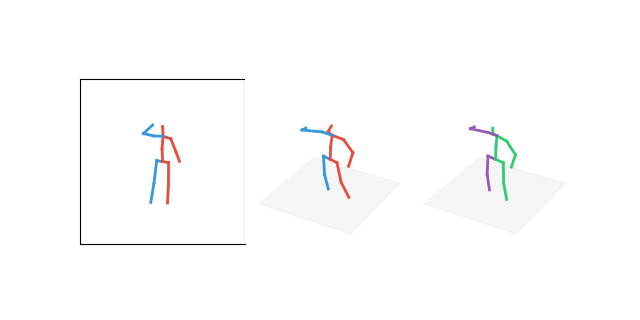}  
    \captionof{figure}{Discussion Pose Estimation Result }
\end{center}

\begin{center}
    \includegraphics[width = .5\textwidth]{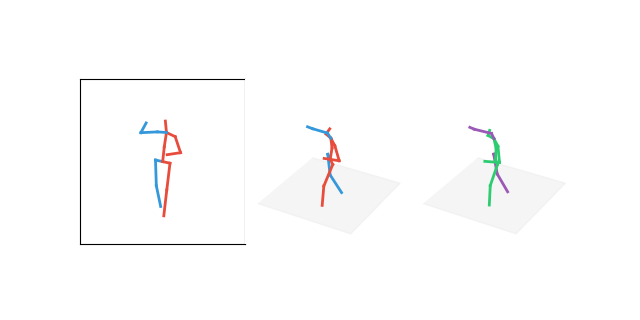}  
    \captionof{figure}{Greeting Pose Estimation Result}
\end{center}

\begin{center}
    \includegraphics[width = .5\textwidth]{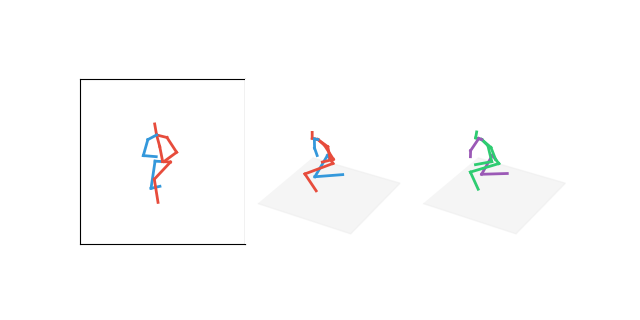}  
    \captionof{figure}{Posing Pose Estimation Result}
\end{center}

\begin{center}
    \includegraphics[width = .5\textwidth]{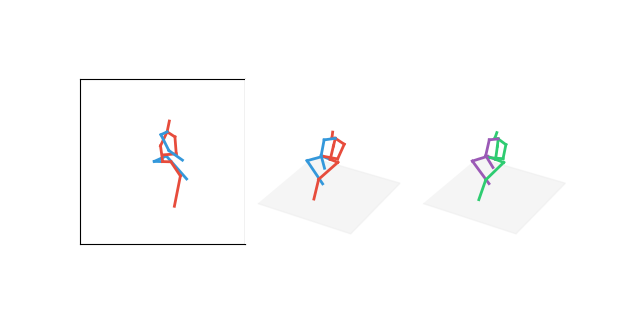}  
    \captionof{figure}{Smoking Pose Estimation Result}
\end{center}

\section{Discussion}
Judging from the Table 1 and Table 2 presented above, the improvement in version 1 showed that these additional layers appear to have had a positive effect on the capability of estimating the majority of the tested poses, while the even more satisfactory results in version 2 suggest that the modifications of these training parameters that we previously mentioned improved the model's performance even further.

Despite the undesired result of version 3, it does not mean that the refinements were made in vain. The reason behind such a result is because the Weighted Mean Squared Error we used to train the human pose data was not applied to the data testing process due to the limited time for the study. In this case, version 3 went through the testing with all the joints with the same weights (MPJPE), while the different weights that should be assigned to the joints based on their importance would make a tremendous difference in the outcome. In fact, we did have some iterations where we reduced the difference of joint weights and they produced better MPJPE results, but their visualized prediction results were not as obvious as the version 3 predicted, as shown above in Figure 10 to 14. Therefore, we did not use them in this report as version 3.

In addition, upon analyzing the results, it can be found that there are both similarities and differences in the MPJPE values of all the versions across the different poses. For example, in the $Eating (eat)$ pose, all four versions had similar MPJPE values, with no significant difference between them. On the other hand, in the $Discussion (discuss.)$ and the $Greeting (greet)$ pose, version 1 and version 2 outperformed the original version greatly, while version 3 had a slightly higher MPJPE value. It suggests that the specific characteristics of the pose may also have impacts on their MPJPE results and that is something our future study can investigate.

\section{Conclusion \& Future Work}
In conclusion, in this paper we managed to propose an innovative advanced baseline model for 3D human pose estimation. After quantitative comparisons with the one proposed in \cite{https://doi.org/10.48550/arxiv.1705.03098}, we can conclude that we have successfully optimized the network architecture as well as some important training parameters to achieve noticeably better results. In addition, we have also proposed a weighted loss function that gives different weights to different human joints, which can enjoy better performance in actual usages.

Future work can also further look into the importance rank of each human joint, and give different weights accordingly to fit the model into more practical usage. A more suitable evaluation model can also be developed accordingly. In addition, attempts to add convolution layers and pooling layers to the training model can also be made to see if they can further improve the results while keeping the model lightweight.



%

%

\bibliographystyle{IEEEtran}
\bibliography{cite}

\end{document}